\documentclass[runningheads]{llncs}

\usepackage{eccv}

\usepackage{eccvabbrv}

\usepackage{graphicx}
\usepackage{booktabs}

\usepackage[accsupp]{axessibility}  

\usepackage[export]{adjustbox}  
\usepackage{multirow}
\usepackage{xcolor}
\usepackage{booktabs}
\usepackage{enumitem}
\usepackage{pifont}

\usepackage{graphicx}
\usepackage{multirow}
\usepackage{rotating}
\usepackage{booktabs}

\usepackage{graphicx}
\usepackage{booktabs}
\usepackage{pifont}

\newcommand{\cmark}{\ding{51}} 
\newcommand{\xxmark}{\ding{55}} 
\newcommand{\xmark}{--}

\definecolor{red}{rgb}{0.95,0.4,0.4}
\definecolor{green}{rgb}{0.55,1.0,0.55}
\definecolor{lightgreen}{rgb}{0.75,1.0,0.75}
\definecolor{blue}{rgb}{0.4,0.4,0.95}
\definecolor{darkblue}{rgb}{0,0,0.8}
\definecolor{darkred}{rgb}{0.8,0,0}
\definecolor{darkgreen}{rgb}{0,0.5,0}
\definecolor{grey}{rgb}{0.6,0.6,0.6}
\definecolor{amber}{RGB}{255,210,43}

\usepackage{hyperref}

\usepackage{orcidlink}

\begin{document}

\title{Towards Zero-Shot Camera Trap Image Categorization} 


\author{Jiří Vyskočil\inst{1}\orcidlink{0000-0002-6443-2051} \and
Lukas Picek\inst{1,2}\orcidlink{0000-0002-6041-9722}}

\authorrunning{J.~Vyskočil and L.~Picek}

\institute{Faculty of Applied Sciences, University of West Bohemia, Pilsen, Czechia \and
Inria, LIRMM, Université de Montpellier, CNRS, Montpellier, France\\}

\maketitle

\begin{abstract} 
This paper describes the search for an alternative approach to the automatic categorization of camera trap images. First, we benchmark state-of-the-art classifiers using a single model for all images. Next, we evaluate methods combining MegaDetector with one or more classifiers and Segment Anything to assess their impact on reducing location-specific overfitting. Last, we propose and test two approaches using large language and foundational models, such as DINOv2, BioCLIP, BLIP, and ChatGPT, in a zero-shot scenario. Evaluation carried out on two publicly available datasets (WCT from New Zealand, CCT20 from the Southwestern US) and a private dataset (CEF from Central Europe) revealed that combining MegaDetector with two separate classifiers achieves the highest accuracy. This approach reduced the relative error of a single BEiTV2 classifier by approximately 42\% on CCT20, 48\% on CEF, and 75\% on WCT. Besides, as the background is removed, the error in terms of accuracy in new locations is reduced to half. The proposed zero-shot pipeline based on DINOv2 and FAISS achieved competitive results (1.0\% and 4.7\% smaller on CCT20, and CEF, respectively), which highlights the potential of zero-shot approaches for camera trap image categorization.
\end{abstract}


\section{Introduction} \label{sec:intro}
Camera traps are valuable assets in ecological research. They are commonly used to estimate wildlife populations, species distribution, and their interactions\cite{blount2021covid,o2011camera,rowcliffe2008estimating,vidal2021perspectives}.
In many cases, the data (i.e., images) are still processed manually, which is extremely time-consuming, given the relatively high number of operated camera traps and their continuous data flow. Therefore, a concerted effort is being made to automate this process using machine learning and computer vision\cite{norouzzadeh2021deep,willi2019identifying}. 
While some studies have achieved human-level performance\cite{valan2020mastering,willi2019identifying}, some challenges persist. For instance, models trained on a set of locations perform poorly in new ones, and models trained in the closed set setting must be re-trained. The same applies to new and unseen species, where adaptability is crucial, as it ensures that if a new, rare animal species appears in the monitored area, the classifier can recognize it.

In this paper, we test how existing foundational models perform in automatic camera trap image categorization and if they allow to overcome the above-mentioned problems.
First, we evaluate state-of-the-art CNN- and Transformer-based classification architectures on three datasets (i.e., WCT\cite{anton2018monitoring} from New Zealand, CCT20\cite{beery2018recognition} from California, and CEF from Central Europe\footnote{Due to animal safety concerns this dataset is available only after signing a Non-Disclosure Agreement (NDA).}) from different continents and with different species.
Second, we use MegaDetector (MD)\cite{Microsoft2023MegaDetector} and Segment Anything (SAM)\cite{kirillov2023segany} models for zero-shot detection and segmentation and test how these models improve classification performance and how they can help mitigate the issue of overfitting to the location/background.
At last, we propose and test two approaches using large language and foundational models, such as DINOv2\cite{oquab2023dinov2,darcet2023vision}, BioCLIP\cite{stevens2024bioclip}, BLIP\cite{li2022blip}, and ChatGPT\cite{FreeGPT}, in a zero-shot scenario that perform similarly good in \textit{seen} and \textit{new locations} as the best approach based on supervised learning. 

\section{Related Work} \label{sec:relwork}
In the past, camera trap image categorization was often done manually, which was neither effective nor fast\cite{swanson2015snapshot,fegraus2011data}. In response, ecologists and the machine learning community naturally shifted their interest in developing novel methods based on machine learning and computer vision.

Pioneer studies in automatic camera trap image categorization\cite{yu2013automated,harris2010automatic} were based on local features (e.g., SIFT, SURF) or early adoptions of neural network classifiers. With the fast progress in machine learning, especially CNNs, many studies\cite{Willi2018,miao2019insights,NorouzzadehE5716,willi2019identifying} focused on fine-tuning standardized architectures for image categorization such as ResNet and VGG.
Recent work\cite{cunha2023bag,henrich2024semi,norouzzadeh2021deep,schneiderrecognizing,shepley2021automated} employs a two-step process involving object detection\footnote{Usually using large pre-trained models, e.g., MegaDetector\cite{Microsoft2023MegaDetector}.} followed by classification. This approach primarily mitigates information loss caused by resizing images to fit the expected classifier input and consequentially reduces overfitting to training locations. Additionally, detecting the animals has clear advantages, as it enables handling multi-species presence and counting the animals present, in addition to categorization.

Even though there is a long track record of methods for camera trap image categorization, most existing approaches share the following drawbacks: 
(i) Methods that do not use object detection to crop the animal before classification tend to overfit to the background, resulting in poor performance on new and \textit{unseen} locations. Additionally, even with the cropped animal, some background pixels are present and, therefore, can result in overfitting. 
(ii) All available pre-trained models, trained on a closed set of categories (i.e., species), cannot be effectively deployed in different climates or continents where species not present in the training data are naturally present. (iii) Existing datasets lack standardization, and available categories are usually on a different taxonomical level, i.e., species, genus, family.

\newpage
\section{Datasets} \label{sec:datasets}
While selecting the datasets for our experiments, we considered the different geographical locations and species. Therefore, we evaluated our experiments on three datasets that originate from North America (Caltech Camera Traps-20), Oceania (Wellington Camera Traps), and Europe (Central Europe Fauna-22).

We use CCT20 as the primary subject in our ablations due to its smaller scale (i.e., faster training) and wide usage. To verify the outcomes in different settings and geographic locations, we use the WCT and CEF22 datasets. Below, we briefly describe each dataset, with statistics provided in Table~\ref{tab:datasets}.

\vspace{0.3cm}
\noindent\textit{\textbf{Note}: Typically, there is only one animal per image, but in rare cases where multiple species appear, we remove those images from the evaluation. Besides and similarly as in the original work\cite{beery2018recognition}, we also remove images without animals.}

\begin{table}[!h]
\vspace{-0.25cm}
    \setlength{\tabcolsep}{4.25pt}
    \centering
    \caption{Statisticts for selected camera trapping datasets. $^\dagger$Denotes custom split.}
    \vspace{-0.2cm}
    \begin{tabular}{@{}l|cc|rrr@{}}
    \toprule
    Dataset                & Boxes  & Categories & Training & Validation & Test       \\
    \midrule
    Caltech Camera Traps\cite{beery2018recognition}          & \cmark & 16     & ~~13,553 &    ~~5,209 & 39,102       \\
    (our) Central Europe Fauna                                & \xmark & 32     & ~~92,603 &     26,479 & ~~~~ -- ~~~~ \\
    Wellington Camera Traps$^\dagger$\cite{anton2018monitoring} & \xmark & 17     & 140,035  &     35,009 & 95,406       \\
    \bottomrule
    \end{tabular}
    \label{tab:datasets}
    \vspace{-0.4cm}
\end{table}

The \textbf{Caltech Camera Traps-20 (CCT20)}\cite{beery2018recognition} dataset consists of images captured from 20 locations in the Southwestern United States. This dataset includes approximately 58,000 images, covering 15 animal categories as well as one "empty" image category. The data are neither balanced nor filtered, reflecting a more natural distribution of species across both time and location.
To allow evaluation of location/environment regularization, the validation set is divided into two subsets: \textit{cis} and \textit{trans}. 
The \textit{cis} subset includes only locations included in the training set, while the \textit{trans} subset includes just new locations. 

The \textbf{Wellington Camera Traps (WCT)}\cite{anton2018monitoring} dataset originates from 187 locations in New Zealand and contains around 270,000 images across 16 categories, plus one category for empty and unclassifiable images.
Since the dataset is not pre-split into train, validation, and test subsets, we allocated 1/3 of the locations for testing and the remaining 2/3 for development, which includes both training and validation. Within the development set, the images are randomly divided in an 80/20 ratio for training and validation, respectively. This approach ensures a balanced distribution for model development and evaluation.

Unlike the previous two datasets, the \textbf{Central Europe Fauna-22 (CEF22)} is a private dataset originating from three different trapping projects and multiple distinct regions within Central Europe. The dataset consists of approximately 120,000 images and includes 32 species categories\footnote{Except two taxon labels (e.g., Aves and Martes) data are labeled with species labels.}. Notably, all empty images were prefiltered and are not included in the dataset. This curated approach focuses exclusively on images containing identifiable species, enhancing the dataset's utility for training and evaluation purposes.

\section{Baseline Performance} \label{sec:experiments}
\begin{figure}[h]
    \centering
    \includegraphics[width=0.95\linewidth]{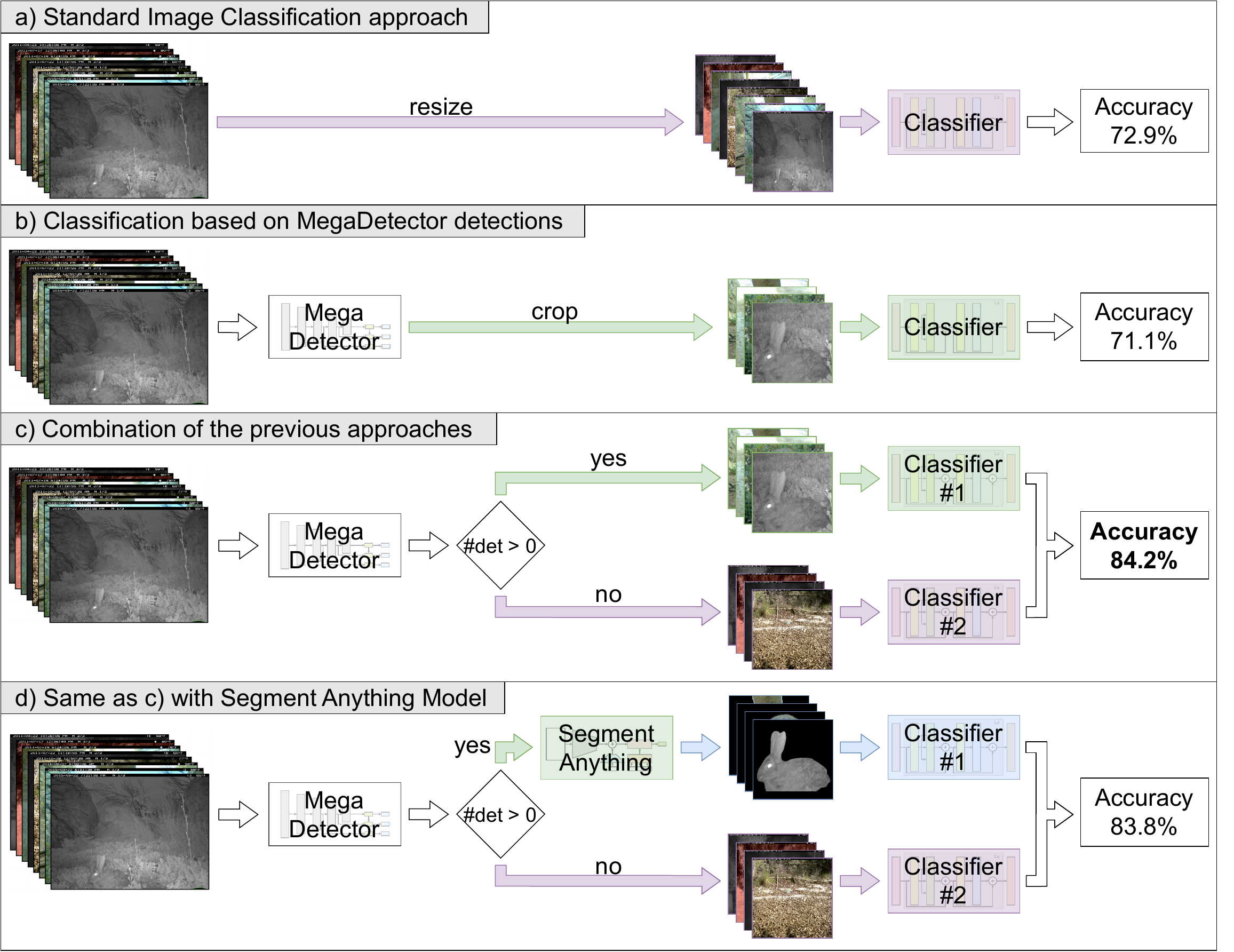}
    \caption{\textbf{Four baseline approaches illustration.} 
    (a) Standard image classification approach where images are resized and fed to a trained image classifier.
    (b) A more recent two-stage approach with object detection before the classification. This approach can suffer from missing reject options for images with no detection.
    (c) With two image classifiers, the problem of (b) can be easily mitigated.
    (d) Excluding background pixels using SAM could help to prevent overfitting to the location.} 
    \label{fig:illustration}
\end{figure}

To the best of our knowledge, there is no standardized benchmark for camera trap categorization (including the three selected datasets); therefore, we provide a few baseline experiments that evaluate state-of-the-art CNN- and Transformer-based architectures for image classification.
This section presents multiple experiments and ablations that focus on benchmarking performance for CNN- and transformer-based classifiers. In the following section, these models are further enhanced with existing foundational models in a zero-shot detection and segmentation context. 
In Figure \ref{fig:illustration}, we illustrate four designed approaches. \\

\noindent\textbf{Evaluation protocol:} In all the experiments and ablations presented in this section, we use the Top1 and Top3 accuracy and macro averaged F1 score (F1$_m$) as the evaluation metrics. The evaluation is primarily carried out on CCT20 validation subsets \mbox{(i.e., cis\_val + trans\_val)} and further tested on CEF22 and WCT datasets to verify the results transferability.

\noindent\textbf{Benchmarking image classifiers:}
Based on the results reported in general image classification\cite{cai2022efficientvit,peng2022beitv2} and species classification\cite{picek2022danish,picek2022plant}, we primarily focus on transformers, but we add ConvNext and ResNext into the mix to compare it for camera trap image analysis. \vspace{-0.2cm}\\

\noindent\textbf{Experiment settings:} We use pre-trained models from the timm python library\cite{rw2019timm}, version 0.9.12. We train all models using SGD optimizer and the momentum of 0.9\cite{qian1999momentum} for 40 epochs with a batch size of 64, and the cosine scheduler, which decreases the learning rate from 1e$^{-3}$ to 2.54e$^{-6}$, and the following augmentations during the training: \textit{RandomResizeCrop} with a scale from 0.8 to 1.0, and RandAugment\cite{cubuk2020randaugment} with a magnitude of 0.2.\footnote{Selected as sub-optimal for all selected models based on our preliminary experiments.} \vspace{-0.2cm}\\

\noindent\textbf{Results:} Achieved results are reported in Table~\ref{tab:all_cls} and Table~\ref{tab:cls_results}. As expected, transformers and CNNs performed competitively, but transformers achieved better performance in new locations. All models underperformed significantly on the unseen \textit{trans locations}, while the CNNs exhibit a higher score difference between the \textit{trans} and \textit{cis}, indicating that they overfit more on the locations themselves. For the following experiments, we will use just the top 3 architectures, i.e., BEiT, BEiTV2, and EfficientViT.
\begin{table}[h]
\vspace{-0.35cm}
    \setlength{\tabcolsep}{3.5pt}
    \centering
    \caption{
        \textbf{Ablation on model architecture}. We benchmark selected CNN- and transformer-based architectures on the CCT20 val. sets without empty images. The ranking in camera trap image categorization more-less follows the ranking on ImageNet.
    }
    \vspace{-0.2cm}
    \begin{tabular}{@{}lcc|cc|cc|c@{}}
    \toprule
     & \multicolumn{1}{c}{Pre-train} & \multicolumn{1}{c|}{Input} & \multicolumn{2}{c|}{\textit{Cis location}} & \multicolumn{2}{c|}{\textit{Trans location}} & \textit{Both} \\
    Architecture & \multicolumn{1}{c}{checkpoint} & \multicolumn{1}{c|}{size} & ~Top1~ & ~F1$_m$~ & ~Top1~ & ~F1$_m$~ & ~Top1~ \\
    \midrule
    \midrule
    ConvNeXt-Base\cite{liu2022convnet}        & IN22k   & $224^2$ & 84.9 & 68.2 & 36.2 & 20.5 & 65.7 \\ 
    ResNeXt-50\cite{xie2017aggregated}        & IN1k    & $224^2$ & \textbf{86.8} & \textbf{69.7} & 35.8 & 16.4 & 66.7 \\ 
    \midrule
    EfficientViT-B3\cite{cai2022efficientvit} & IN1k    & $224^2$ & 78.9 & 61.2 & 40.9 & 19.4 & 63.9 \\  
    ViT-Base/p16\cite{dosovitskiy2020image}   & IN1k    & $224^2$ & 84.6 & 68.0 & 49.4 & 20.9 & 70.7 \\  
    SwinV2-Base/w16\cite{liu2022swinv2}       & IN1k    & $256^2$ & 85.4 & 66.9 & 49.1 & 21.9 & 71.1 \\  
    Swin-Base/p4-w7\cite{liu2021swin}         & IN22k   & $224^2$ & 83.6 & 67.0 & 52.0 & \underline{24.3} & 71.1 \\  
    BEiT-Base/p16\cite{bao2021beit}           & IN22k   & $224^2$ & 84.4 & 67.2 & \underline{54.4} & 21.1 & 72.6 \\  
    BEiTV2-Base/p16\cite{peng2022beitv2}      & IN22k   & $224^2$ & \underline{85.9} & \underline{68.6} & 53.0 & 21.7 & \underline{72.9} \\  
    EfficientViT-L3\cite{cai2022efficientvit} & IN1k    & $224^2$ & 83.5 & 67.0 & \textbf{60.4} & \textbf{28.6} & \textbf{74.4} \\  
    \bottomrule
    \end{tabular}
    \label{tab:all_cls}
    \vspace{-0.2cm}
\end{table}

\begin{table}[h]
\vspace{-1.0cm}
    \centering
    \setlength{\tabcolsep}{3.5pt}
    \caption{
        \textbf{Transformers performance.} All models achieved a competitive accuracy on all three datasets. However, the BEiTv2 model performed much better on the tail species.
        $^*$fps was measured on NVIDIA A40 (GPU) and AMD EPYC 7543 (CPU).
    }
    \vspace{-0.2cm}
    \begin{tabular}{@{}l|ccc|ccc|ccc|c@{}}
        \toprule
         & \multicolumn{3}{c|}{CCT20 dataset} & \multicolumn{3}{c|}{CEF22 dataset} & \multicolumn{3}{c|}{WCT dataset} &         \\

        Architecture    & Top1 & Top3 & F1$_m$                          & Top1      & Top3      &F1$_m$         & Top1     & Top3      & F1$_m$        & fps$^*$ \\
        \midrule
        BEiT-Base/p16   & 72.6 & \underline{89.7} & 60.1                & 85.3 & 94.5 & \underline{68.4}                 & \textbf{86.5}   & 98.0 & \underline{72.7}   & \underline{331} \\
        BEiTV2-Base/p16 & \underline{72.9} & 89.6 & \underline{61.9}    & \textbf{87.5} & \textbf{95.5} & \textbf{72.2} & \underline{86.0} & \textbf{98.8} & \textbf{75.5}   & \textbf{333} \\
        EfficientViT-L3 & \textbf{74.4} & \textbf{90.8} & \textbf{62.7} & \underline{85.7} & \underline{94.9} & 62.7  & 85.1    & \underline{98.7}    & 67.1   & 295 \\
        \bottomrule
    \end{tabular}
    \label{tab:cls_results}
    \vspace{-0.15cm}
\end{table}

\section{Zero-shot Detection and Segmentation}
\label{subsec:MD_and_SAM}

This section analyzes and provides a qualitative and quantitative evaluation of the potential of two pre-trained foundational models, the MegaDetector (MD)~\cite{Microsoft2023MegaDetector} and the Segment Anything Model (SAM)~\cite{kirillov2023segany}, in processing camera trap images.
Furthermore, we explore decision-making strategies when MD detects no object, including a) considering an image to be empty and b) classifying the whole image. 
At last, leveraging the insights gained from our ablations, we test three state-of-the-art transformer-based architectures in combination with MD and SAM trained on three distinct datasets with different species and origins.  \vspace{-0.2cm}\\

\begin{figure}[b]
    \centering
    \includegraphics[width=0.975\linewidth]{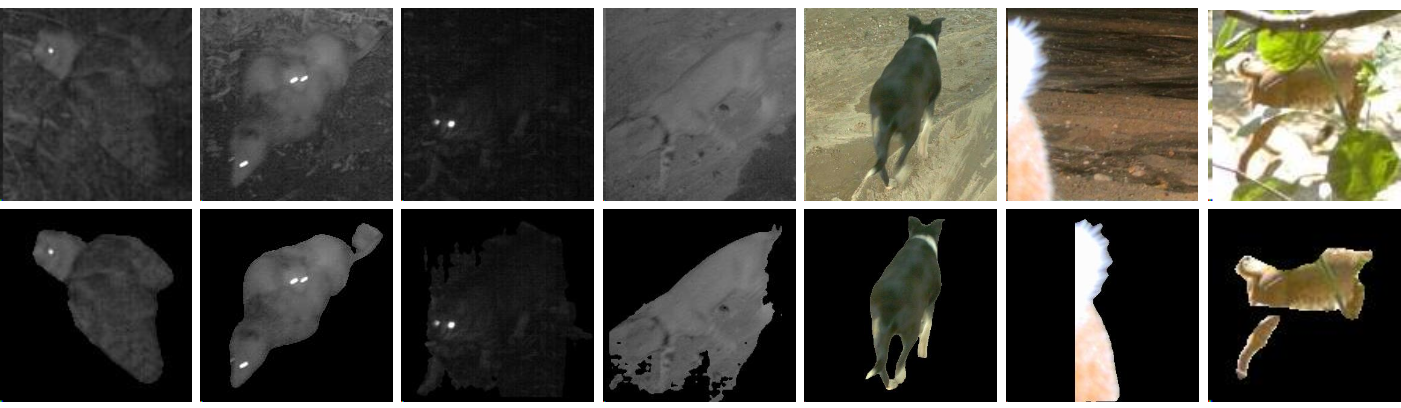}
    \caption{
        \textbf{Zero-shot segmentation with Segment Anything}. Random samples from CCT20 datasets are processed by MegaDetector, and resulting detections are fed into SAM. Even with the poor quality of the data, the zero-shot segmentation performs relatively well across a wide range of species. However, with the infra-red images and the small size of an object, the SAM starts to fail (3$^{rd}$ and $4^{th}$ column from the left).
    }
    \label{fig:crop_examples}
\end{figure}

\noindent\textbf{Image data processing:}
We run MD to obtain detections of objects for all images.
Then crop the detections from the image so that the resulting image has the same width and height to avoid breaking the aspect ratio.
Furthermore, we use the SAM (initialized with the MD's detection) to remove background pixels and move the object to the center. See Figure~\ref{fig:crop_examples} for examples of the detected and segmented objects in the CCT20 dataset. \vspace{-0.2cm}\\

\noindent\textbf{Classification models:}
Since neither MD nor SAM provides class categories, we will use the top three architectures (in terms of accuracy) from the previous ablation. We have four different types of classifiers, each differing in the expected input: (a) full-size images, (b) cropped detections, (c) cropped detections with segmented objects, and (d) both full-size and cropped images. We investigate combinations of these in our experiments and ablation studies.  \vspace{-0.2cm}\\

\noindent\textbf{Building a classifier for inference:}
Once the models are trained, we create a predictor containing two classifiers.
First of all, we apply MD on the image.
If MD finds an object, we use classifier \#1, which is trained to recognize objects from cropped or even segmented images.
If no object is found, we use classifier \#2 on the whole image.
Besides, we perform an experiment in which we use only one classifier on cropped and full-size images.

\subsection{Performance Evaluation} We compare three types of classifiers that differ in their expected input data to explore the benefits of using MD and/or SAM on camera trap images. Each classifier expects one of the following inputs: (a) resized images, (b) images cropped based on detection from MegaDetector, and (c) cropped detections with removed background (we use Segment Anything Model, which is fed with the MegaDetector detections).
The comparison is made on a filtered validation set of the CCT20 dataset that contains only images with objects.
By cropping detections from images, we have improved the Top1 accuracy on average by 10.3\%, 5.2\%, and 10\% on the CCT20, CEF22, and WCT datasets, respectively. The improvement is lower on the dataset with European fauna as it was most likely not used in the training of the MD. Using the segmented animals, the performance decreased by an average of 5.1\% on CCT20, 2.8\% on CEF22, and 0.8\% on WCT. See Table~\ref{tab:MD_and_SAM} for more comprehensive evaluation. \vspace{-0.2cm}\\

\noindent\textbf{Location overfitting:} 
We use the CCT20 \textit{cis} and \textit{trans} subsets to test the hypothesis that image classifiers tend to overfit to background pixels and, therefore, perform poorly in new locations. Following the results listed in Table~\ref{tab:MD_and_SAM}, it is evident that reducing the number of background pixels available for training with both cropping our segmenting the object reduces the 
location overfitting. Besides training the model on cropped objects,  
increases the overall classification performance as more detail is provided. Using MD and "feeding" a classifier with cropped detections still results in around 15\% lower performance in the new locations. However, it is a significant improvement as without the cropping, the resulting performance in \textit{trans} (\textit{unseen}) locations dropped by roughly 30\% in terms of Top1 accuracy, and the error almost doubled.

\begin{table}[h]
 \vspace{-0.25cm}
\setlength{\tabcolsep}{3.25pt}
\caption{
\textbf{Ablations on location overfitting}.
We compare the performance in terms of Top1 Accuracy of three models trained on (a) resized images, (b) cropped animals, and (c) cropped + segmented animals on three datasets.  While (a) shows huge overfitting to the location, the (b) and (c) approaches show much better generalization and robustness to changes in testing location. In general, (b) achieves the best performance.}
    \centering
        \begin{tabular}{l@{\hspace{15pt}}cc@{\hspace{5pt}}|@{\hspace{7pt}}c@{\hspace{7pt}}c@{\hspace{7pt}}c@{\hspace{7pt}}|@{\hspace{7pt}}c@{\hspace{7pt}}c}
        \toprule
                &  &  &  &  &  & \multicolumn{2}{c}{CCT} \\ 
        Architecture        & MD & SAM & CCT & CEF & WCT & \textit{cis} & \textit{trans} \\ 
        \midrule
        BEiT-Base/p16   & \xmark & \xmark & 72.6 & 85.3 & 86.5 & 84.4 & 54.4 \\
        BEiT-Base/p16   & \cmark & \xmark & \textbf{83.9} & \textbf{91.1} & \textbf{96.0} & \textbf{89.6} & \textbf{75.2} \\
        BEiT-Base/p16   & \cmark & \cmark & \underline{80.8} & \underline{88.8} & \underline{95.0} & \underline{87.3} & \underline{70.8} \\
        \midrule
        BEiTV2-Base/p16& \xmark & \xmark & 72.9 & 87.5 & 86.0 & 85.9 & 53.0 \\
        BEiTV2-Base/p16 & \cmark & \xmark & \textbf{84.2} & \textbf{92.2} & \textbf{96.6} & \textbf{90.1} & \underline{75.0} \\
        BEiTV2-Base/p16 & \cmark & \cmark & \underline{83.8} & \underline{89.2} & \underline{95.8} & \underline{89.2} & \textbf{75.5} \\
        \midrule
        EfficientViT-L3 & \xmark & \xmark & 74.4 & 85.7 & 85.1 & 83.5 & 60.4 \\
        EfficientViT-L3 & \cmark & \xmark & \textbf{83.1} & \textbf{90.8} & \textbf{95.0} & \textbf{87.2} & \textbf{76.9} \\
        EfficientViT-L3 & \cmark & \cmark & \underline{81.1} & \underline{87.6} & \underline{94.4} & \underline{85.5} & \underline{74.2} \\
        \bottomrule
        \label{tab:MD_and_SAM}
    \end{tabular}
    \vspace{-0.25cm}
\end{table}



\subsection{Additional Ablations}
\noindent\textbf{Dealing with images without detections:}
Although MD is a powerful foundational model, it has its limitations, such as generating false positives or failing to detect objects altogether. Therefore, we provide an additional ablation that compares two classification methodologies when MD does not detect any objects: (a) using a single classifier for both cropped and full-size images and (b) using separate classifiers for each.

Using a single classifier (a) over the two separate classifiers (b) reduces computational complexity, but it reduces the Top1 accuracy by around 1\% for all three architectures. For example, EfficientViT improves accuracy by 1.4\% with two classifiers, as shown in the Table ~\ref{tab:single_cls_vs_two_cls}. \\

\noindent\textbf{What to do with "empty" images?} A common issue in camera trapping is false sensor activation, leading to saving "empty" images with no animals. In this ablation, we test two approaches to distinguish empty images from those with animals. We use the CCT20 dataset, which includes an "empty" category in the validation and test sets, though not in the training set.

We test the following two approaches:
\begin{enumerate}[label=(\alph*),leftmargin=12mm]
\item Treat images as "empty" when MD detects no object.
\vspace{0.2cm}
\item Generate or add object-free data to the training set and train a second classifier on full-size images. We generate average images based on location, date, and time.
\end{enumerate}

Results in Table~\ref{tab:empty_imgs} show that assuming an image is empty when MD does not detect anything significantly reduces performance by up to 18.5\%, highlighting the need for better strategies in handling empty images.

\begin{table}[h]
\vspace{-0.2cm}
    \setlength{\tabcolsep}{4.5pt}
    \centering
    \begin{minipage}{.4875\linewidth}
        \centering
        \caption{
            \textbf{Classification of images with none MegaDetector detections.}
            Performance of (a) one classifier for cropped images from detections and full-size images and (b) two separated classifiers, one trained on cropped images and the second one on full-size images.
        }
        \begin{tabular}{lc|c}
            \toprule
            Architecture    & Classifier & Top1                             \\
            \midrule
            BEiT-Base/p16   & (a)       & 66.5                           \\
            BEiT-Base/p16   & (b)       & \underline{67.5}\vspace{.5pt}              \\
            \midrule
            BEiTV2-Base/p16 & (a)       & 67.4                           \\
            BEiTV2-Base/p16 & (b)       & \textbf{68.2\vspace{.5pt}}     \\
            \midrule
            EfficientViT-L3  & (a)       & 65.8                           \\
            EfficientViT-L3  & (b)       & 67.2                           \\
            \bottomrule
        \end{tabular}
        \label{tab:single_cls_vs_two_cls}
    \end{minipage}
    \hskip10pt
       \begin{minipage}{.45\linewidth}
        \centering
        \caption{
            \textbf{Dealing with empty images.} We compare a method that trusts the MD to declare the image empty if no object is found with another that uses two classifiers: 1\textsuperscript{st} on a cropped image if an object is detected and 2\textsuperscript{nd} on the full image if not.
        }
        \begin{tabular}{l|@{\hskip 7px}c}
            \toprule
            If no detection, then                & Top1                          \\
            \midrule
            consider it as \textit{empty image}.                  & 49.1                         \\
            use 2\textsuperscript{nd} classifier. & \underline{67.5}                         \\
            \midrule
            consider it as \textit{empty image}.                  & 49.7                         \\
            use 2\textsuperscript{nd} classifier. & \textbf{68.2}                \\
            \midrule
            consider it as \textit{empty image}.                  & 50.1                         \\
            use 2\textsuperscript{nd} classifier. & 67.2                         \\
            \bottomrule
        \end{tabular}
        \label{tab:empty_imgs}
    \end{minipage}
\end{table}

\section{Zero-shot Classification}

\begin{figure}[h!]
\vspace{-0.25cm}
    \centering
    \includegraphics[width=0.96\linewidth]{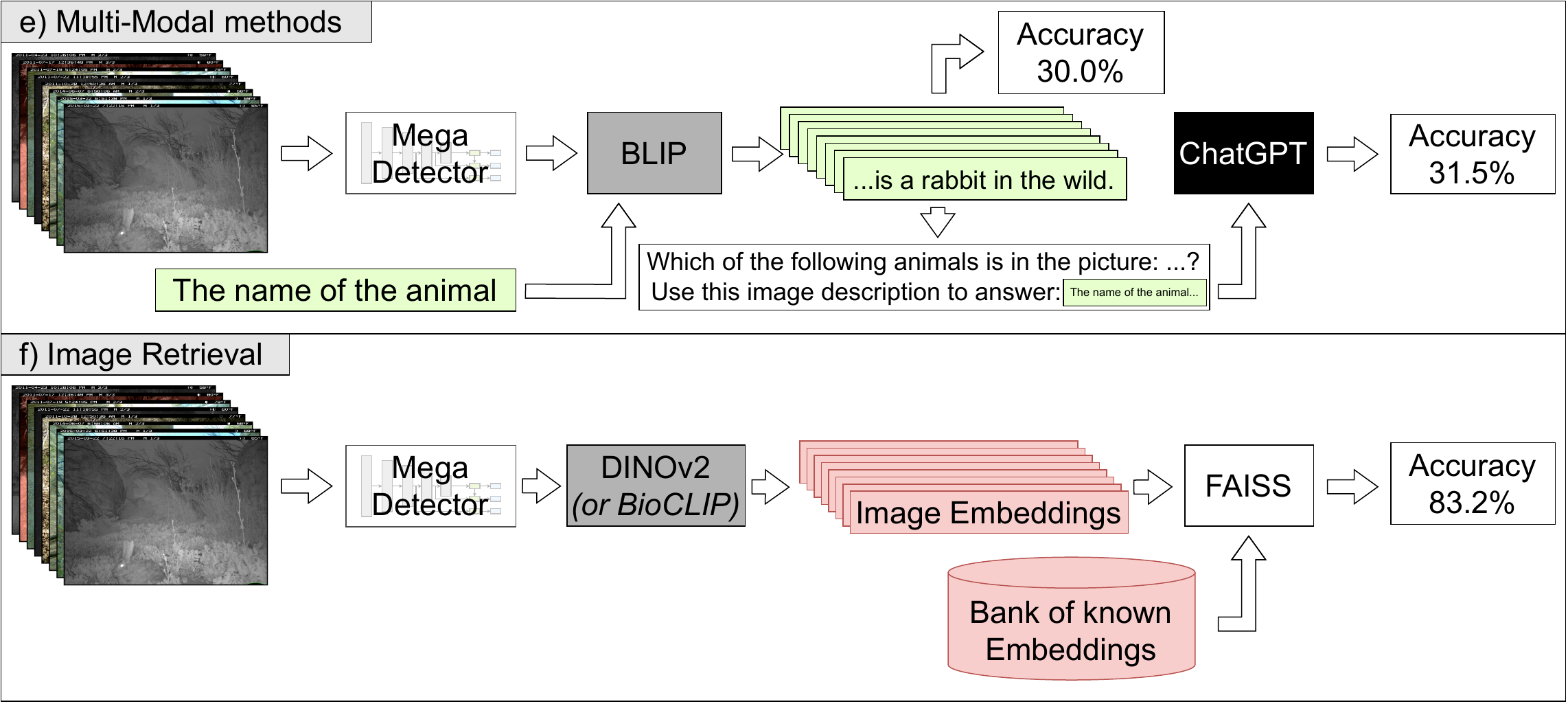}
    \caption{\textbf{Zero-Shot approaches.} 
    (e) Multi-modal methods extract image info. based on given textual and image prompts.
    (f) Embeddings are generated from the training set, and during inference, selective search finds similar images from the database.}
    \label{fig:zero_shot_cls}
    \vspace{-0.75cm}
\end{figure}

In addition to the standard classification approach with trained closed-set model for camera trap image categorization, and following the success of LLMs, we further explore the capabilities of existing foundational models for multi-modal processing (e.g., BLIP\cite{li2022blip} and ChatGPT\cite{FreeGPT}) and image retrieval using BioCLIP\cite{stevens2024bioclip} and DINOv2\cite{oquab2023dinov2}. For "classification" based on image retrieval, we use the FAISS library\cite{johnson2019billion,douze2024faiss} developed for efficient similarity search and clustering of deep embeddings. Both evaluated pipelines are illustrated in Figure~\ref{fig:zero_shot_cls}. \\

\noindent\textit{\textbf{Note:} We are aware of the limited reproducibility when using ChatGPT, but one of the goals of this work is to fully explore what are the current LLMs capabilities in scenarios related to ecology compared to standardly used methods.}

\subsection{Large Language and Multi-Modal Models}
We use an image captioning model, BLIP, that provides a textual output to a prompt consisting of text and image. 
To enhance BLIP's focus on animals, we condition various image captions by providing text that includes keywords such as "what," "species," "animal," etc. Furthermore, we use ChatGPT (based on GPT-4) as a filter to identify animals from the given options and image description. We consider the image to be empty if none or more than one of the possible categories is included in the response. \\

\noindent\textbf{BLIP}~\cite{li2022blip}: We test how the pre-trained multi-modal image captioning model's (i.e., BLIP) performance depends on different conditional captioning inputs (e.g., \textit{"The species of the animal is..."}, or\,\textit{"The name of the animal..."}).  
Overall the performance fluctuated inconsistently across the datasets and "prompts," and conditioning the output with no captions can yield relatively good scores. 
Using \textit{"A running..."} as captioning yielded the best results on the CEF22 datasets; however, slightly underperformed on the CCT20 and WCT. Still, this captioning seems to be a cross-dataset sub-optimum. 
Interestingly, when the model fails to recognize an animal, it often defaults to “a bear in the wild," even if no bear is present in the dataset, or to phrases like “is not visible”/“is on the camera screen.” For detailed performance, see Table~\ref{tab:BLIP}. Compared to baseline performance with approach (a) and Top1 accuracy of BEiTV2-Base/p16 (i.e., 84.2, 92.2, and 96.6 on CCT20, CEF22, and WCT respectively), BLIP underperformed badly, achieving roughly half of the accuracy.

\begin{table}[h]
\vspace{-0.25cm}
    \centering
    \setlength{\tabcolsep}{3.5pt}
    \caption{\textbf{Ablation on "captioning prompting" of BLIP}.
    }
    \begin{tabular}{l|ccc}
        \toprule
        Conditional image captioning          & CCT20         & CEF22         & WCT  \\
        \midrule
        \textit{--}                           & 21.7          & 32.6          & \underline{30.1} \\ 
        \midrule
        \textit{The picture shows a cute ... }     & 14.7          & 35.6          & 21.8 \\ 
        \textit{I see cute ... }                  & 19.0          & 36.6          & \textbf{30.2} \\ 
        \textit{The species of the animal is ... } & 20.0          & {\hspace{+5.5px}}2.5 & 18.0 \\ 
        \textit{The animal in the picture is ... } & 20.1          & {\hspace{+5.5px}}1.9 & 16.8 \\ 
        \textit{A running ... }                    & 21.5          & \textbf{40.9} & 27.8 \\ 
        \textit{A peeking ... }                   & 22.5          & \underline{38.1}          & 29.2 \\ 
        \textit{This animal is called ... }        & \underline{24.4}          & 28.3          & 24.9 \\ 
        \textit{The name of the animal ... }      & \textbf{24.9} & 37.3          & 25.7 \\ 
        \bottomrule
    \end{tabular}
    \label{tab:BLIP}
\vspace{-0.25cm}
\end{table}

\noindent\textbf{Fine-tuning BLIP captions with ChatGPT:}
ChatGPT has gained popularity due to its suitability across diverse topics and its ability to engage in human-like conversations.
However, its capabilities in camera trap image categorization have not been explored yet. Since the ChatGPT API only allows textual inputs, we use it to further process the BLIP's outputs given the image and sub-optimal conditional captioning and put it as an input into the GPT-4 based model\cite{FreeGPT}. The proposed textual input for the ChatGPT is as follows:
\begin{center}
    \textit{Write a one-word answer to this question:
    "Which of the following animals is in the picture:
    \textcolor{blue}{\textless list of categories\textgreater}?"
    Consider this image description in the answer:
    \textcolor{teal}{\textless BLIP generated caption\textgreater}.}
\end{center}
Where \textit{\textcolor{blue}{\textless list of categories\textgreater}} are all targets without "empty" category, separated with commas and \textit{\textcolor{teal}{\textless BLIP generated caption\textgreater}} is BLIP's output. 

Based on our experiment, ChatGPT can slightly improve the performance of BLIP, with an improvement of approximately 1.5\% on the CCT20 dataset. Our observations indicate that, in some cases, ChatGPT adjusts the final prediction by selecting a class from the options provided. However, more frequently, ChatGPT ignores the options and instead propagates the species identified in the caption generated by BLIP, even if it is not among the provided options (e.g., “bear”). Furthermore, ChatGPT tends to list the options from which it claims to have selected its answer, even when the chosen answer is absent from the list.

\subsection{Retrieval-based Classification}
\begin{figure}[h]
\vspace{-0.5cm}
    \centering
    \setlength\tabcolsep{1pt}
    \begin{tabular}{c@{\hspace{5pt}}|@{\hspace{5pt}}ccc}
        \textbf{Full-size image} & {\textbf{Top1}} & {\textbf{Top2}} & {\textbf{Top3}} \\ 
        \includegraphics[height=2.0cm]{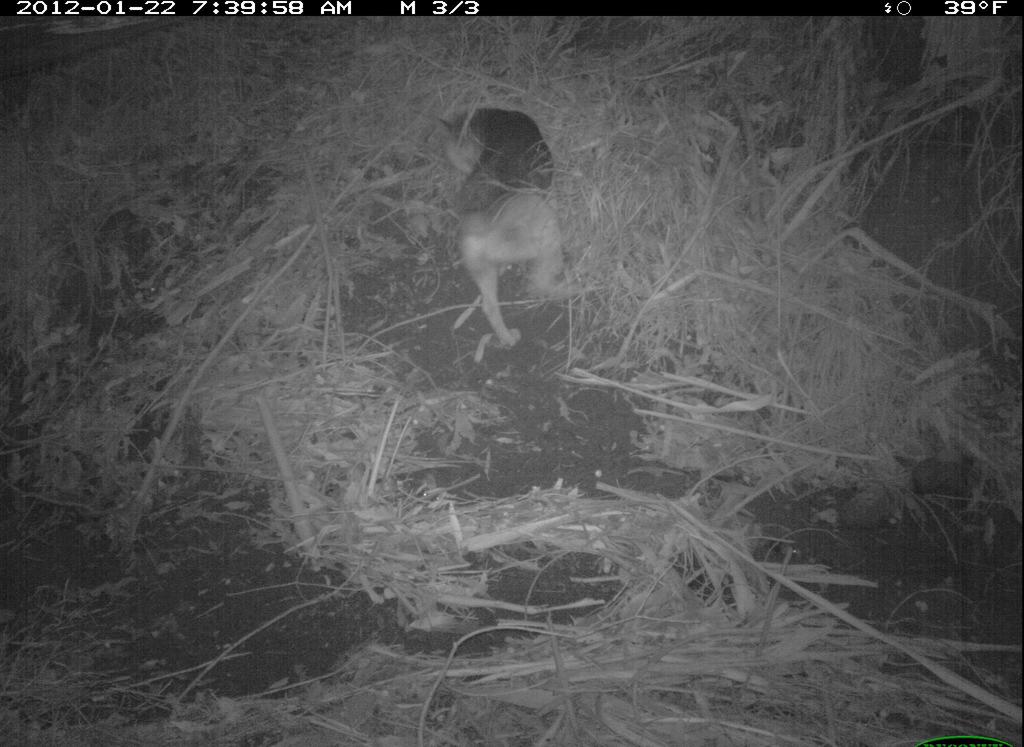} 
        & \includegraphics[height=2.0cm]{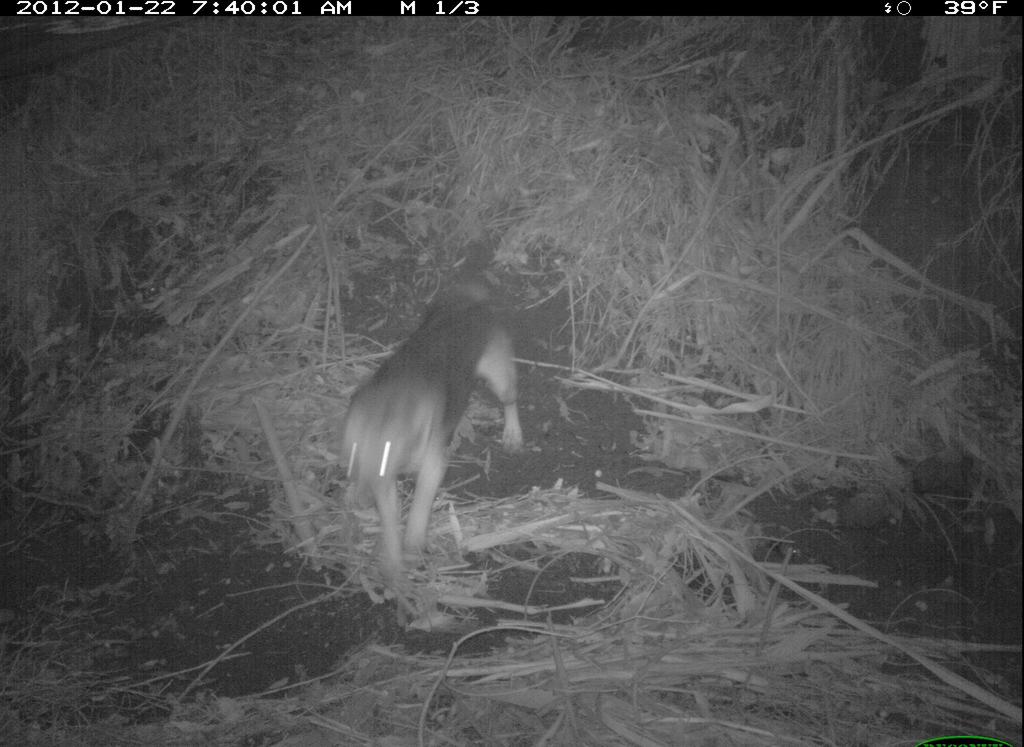} 
        & \includegraphics[height=2.0cm]{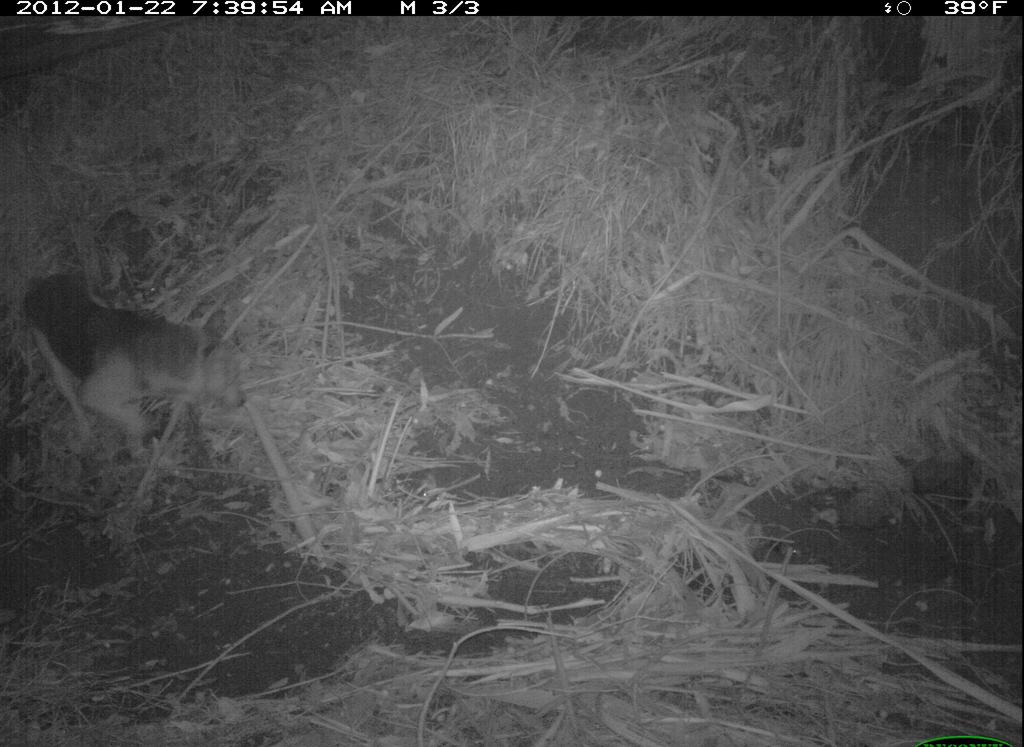} 
        & \includegraphics[height=2.0cm]{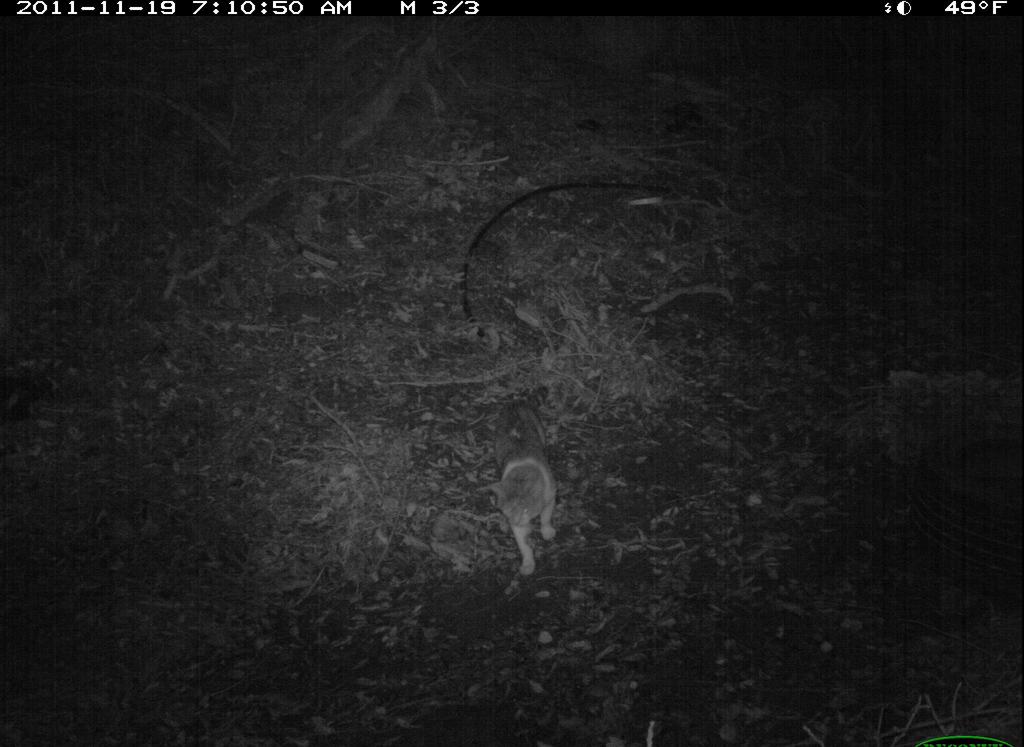} 
        \\
        \raisebox{-3pt}{\textbf{Cropped object}}  & \raisebox{3pt}{dog  \color{darkgreen} \cmark} & \raisebox{3pt}{dog  \color{darkgreen} \cmark} & \raisebox{3pt}{cat  \color{darkred} \xxmark}  \\  [3pt]       
          \includegraphics[height=2.5cm]{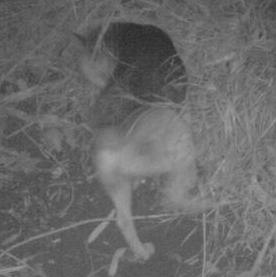} 
        & \includegraphics[height=2.5cm]{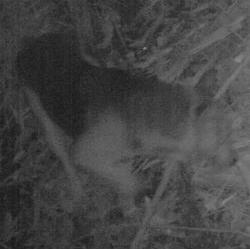} 
        & \includegraphics[height=2.5cm]{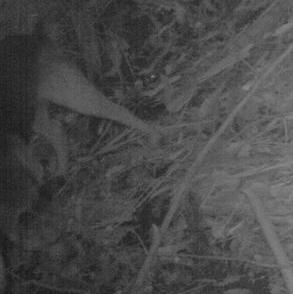} 
        & \includegraphics[height=2.5cm]{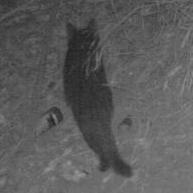} 
        \\       
        \raisebox{-3pt}{\textbf{Segmented object}} & \raisebox{3pt}{dog  \color{darkgreen} \cmark} & \raisebox{3pt}{dog  \color{darkgreen} \cmark} & \raisebox{3pt}{cat  \color{darkred} \xxmark}   \\ [3pt]
        \includegraphics[height=2.5cm]{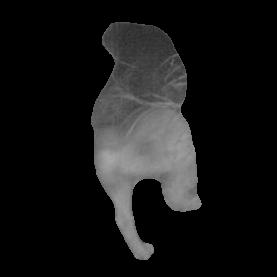} 
        & \includegraphics[height=2.5cm]{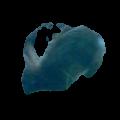} 
        & \includegraphics[height=2.5cm]{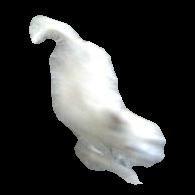} 
        & \includegraphics[height=2.5cm]{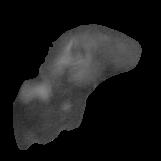} 
        \\
         & \raisebox{3pt}{rabbit  \color{darkred} \xxmark} & \raisebox{3pt}{dog  \color{darkgreen} \cmark} & \raisebox{3pt}{skunk  \color{darkred} \xxmark}   \\ 
    \end{tabular}
    \vspace{-5pt}
    \caption{
        \textbf{Top3 closest images to given inputs using DINOv2$_G$ embeddings.}
    }
    \label{fig:top3_examples}
\end{figure}

In this section, we test the suitability of existing pre-trained foundational models (e.g., BioCLIP\cite{stevens2024bioclip} and DINOv2\cite{oquab2023dinov2,darcet2023vision}) for zero-shot camera trap image categorization in a retrieval-like setting. Zero-shot classification of camera trap images can solve some of the major drawbacks of the existing approaches as it allows the classification of images into categories not seen during training by retrieving and comparing them against a set of predefined class examples. This means that the database can be extended by new species at any time without re-training the model. Furthermore, this approach allows simple updates of the class reference samples, so weekly labeled data and data with no additional value might be easily filtered out or replaced. On the other hand, after the inference, one more step, i.e., similarity search, is needed. For example predictions, see Figure \ref{fig:top3_examples}.\\

\noindent\textbf{Method:}
First, we feed forward all the training images into the image encoders and store image embeddings into the \textit{database}. Second, we generate embeddings for all \textit{query} data the same way. At last, we use FAISS library\cite{johnson2019billion,douze2024faiss} to perform an efficient similarity search. 
In both cases, the images are resized to 256$\times$256 and center-cropped to 224$\times$224. As a baseline, we use the standard L2 distance without feature vector normalization and k-NN classifier with k = 1 to select the species prediction.

\noindent\textbf{Results:}
Overall, the best-performing model was the DINOv2$_G$, which achieved an outstanding performance comparable to models trained with full supervision. More precisely, the DINOv2$_G$ achieved a Top1 accuracy of 83.2\% and 87.5\% on CCT and CEF datasets, respectively. Interestingly, the newer and "\textit{supposed-to-be better}" DINOv2$_B$+reg\cite{darcet2023vision} slightly underperformed the vanilla DINOv2$_B$ and the BioCLIP$_B$ heavily underperformed both DINO models.

As in previous experiments, cropping detected animals was also beneficial for image retrieval, and the segmentation seems to be unnecessary. In the scenario with just a single classifier, the DINOv2$_G$ performed similarly on the CEF dataset and slightly better on the CCT dataset.
A more comprehensive qualitative and quantitative evaluation is listed in Table~\ref{tab:DINOv2}, and Figure~\ref{fig:top3_examples}.

\begin{table}[h]
\vspace{-5pt}
    \centering
    \caption{
        \textbf{Camera trap image categorization in retrieval-like setting}. 
        We compare BioCLIP, DINOv2\cite{oquab2023dinov2}, and DINOv2+reg\cite{darcet2023vision}, on the CCT20 and CEF20 datasets in terms of Top1 accuracy. To allow direct comparison with the \textit{standard} approaches (i.e., (a), (b), and (c) in Figure \ref{fig:illustration}), we add scores achieved with BEiTv2 classifiers explicitly trained to classify camera trap images on those datasets.
    }
    \setlength{\tabcolsep}{3.5pt}
    \begin{tabular}{ccc|ccc|c|c}
        \toprule
                                                       & MD     & SAM    & BioCLIP$_B$ & DINOv2$_B$+reg & DINOv2$_B$ & DINOv2$_G$ & \textit{BEiTv2$_B$} \\ \midrule
        \multirow{3}{*}{\rotatebox[origin=c]{90}{CCT}} & --     & --     & 60.5      & 65.5         & 67.8     & 74.4             & {72.9}           \\
                                                       & \cmark & --     & 65.4      & 77.5         & 77.8     & \underline{83.2} & \textbf{{84.2}}  \\
                                                       & \cmark & \cmark & 62.7      & 71.8         & 70.2     & 75.6             & {83.8}           \\ \midrule
        \multirow{3}{*}{\rotatebox[origin=c]{90}{CEF}} & --     & --     & 70.1      & 80.9         & 81.5     & 87.1             & 87.5             \\
                                                       & \cmark & --     & 69.9      & 82.5         & 82.6     & \underline{87.5} & \textbf{92.2}    \\ 
                                                       & \cmark & \cmark & 61.6      & 74.2         & 75.3     & 81.1             & \underline{89.2} \\
\bottomrule
    \end{tabular}
    \label{tab:DINOv2}
    \vspace{-5pt}
\end{table}

\noindent\textbf{Ablation on location generalization}. 
Given that image retrieval-based classifiers demonstrated comparable results to fine-tuned models, we performed an additional ablation on three African datasets\footnote{We remove empty images and compare models without using MD or SAM.} from Snapshot Safari~\cite{pardo2021snapshot}: Enonkishu (ENO), Kgalagadi (KGA), and Kruger (KRU). As these datasets are not split into development and testing subsets, we use images from the first $x$ locations\footnote{We use the first 8, 15, and 26 locations for the ENO, KGA, and KRU datasets, resulting in 20\%, 31\%, and 29\% of the total images being used for testing, respectively.} to build the database and the remaining locations for testing. In other words, the results reported in Table~\ref{tab:other_datasets} for ENO, KGA, and KRU are based solely on images from unseen locations.

Retrieval-based methods achieve results comparable to, and in some cases exceeding, the BEiTv2 fine-tuned models. This is particularly evident on the African datasets, which contain relatively limited training data for classification (3,929, 662, and 1,312 images for ENO, KGA, and KRU, respectively), where model training completely failed for KRU. Considering that the performance of retrieval-based methods can be easily improved by simply adding new data to the database, whereas standard image classifiers require model fine-tuning, retrieval-based approaches prove promising results, especially for small datasets.

\begin{table}[t]
\vspace{-0.25cm}
    \centering
    \setlength{\tabcolsep}{7.5pt}
    \caption{\textbf{Results of image retrieval-based classifier on additional datasets.}
    }
    \begin{tabular}{l|cccccc}
        \toprule
        Model       & CCT              & CEF              & WCT              & ENO              & KGA              & KRU              \\
        \midrule
        BioCLIP$_B$ & 60.5             & 70.1             & 80.3             & 44.3             & 72.8             & \underline{55.7} \\
        DINOv2$_G$  & \textbf{74.4}    & \underline{87.1} & \underline{83.6} & \textbf{65.9}    & \underline{79.9} & \textbf{79.6}    \\
        \midrule
        \textit{BEiTv2$_B$}  & \underline{72.9} & \textbf{87.5}    & \textbf{86.0}    & \underline{53.5} & \textbf{83.0}    & 41.1             \\
        \bottomrule
    \end{tabular}
    \label{tab:other_datasets}
\vspace{-0.25cm}
\end{table}

\vspace{0.4cm}
\noindent\textbf{Ablation on matching strategy}. In this experiment, we compare two 
similarity measures ($L_2$ norm and cosine similarity) and various $k$ values for the k-NN classifier over DINOv2$_G$ features on five distinct datasets.
The results are inconclusive as the best-performing approach is different for each dataset. Therefore, we recommend testing the best approach before using it on your dataset. For further details, refer to Table~\ref{tab:dinov2_settings}.

\begin{table}[h]
    \vspace{-5pt}
    \centering
    \setlength{\tabcolsep}{7.5pt}
    \caption{\textbf{Matching strategy settings ablation.} We compare how different settings affect DINOv2$_G$ retrieval performance using $L_2$ norm and cosine similarity based on class centroids and various $k$ values in the k-NN algorithm.
    }
    \begin{tabular}{l|cc|ccccc}
        \toprule
                &  \multicolumn{2}{c|}{\textit{Cosine similarity}} & \multicolumn{5}{c}{$L_2$ norm} \\ 
        Dataset & 1-NN & 3-NN & 1-NN & 3-NN  & 5-NN  & 10-NN & centr. \\
        \midrule
        CCT  & \textbf{74.4} & 72.7 & \textbf{74.4} & \underline{74.1} & 72.3 & 71.8 & 54.8   \\
        CEF  & 87.0 & \underline{87.7} & 87.1  & \textbf{87.8} & \textbf{87.8} & \underline{87.7} & 74.2   \\
        ENO  & 66.6    & \textbf{68.6} & 65.9 & 67.9 & \underline{68.4} & 65.9 & 54.6 \\
        KGA  & 80.6             & 78.5 & 79.9 & 79.5 & 79.2 & \underline{83.4} & \textbf{86.9} \\
        KRU  & \underline{80.0}    & 79.8 & 79.6 & 76.1 & 77.2 & \textbf{80.7} & 74.6 \\
        \bottomrule
    \end{tabular}
    \label{tab:dinov2_settings}
\vspace{-0.4cm}
\end{table}

\section{Conclusion and Perspectives}

This paper presents a comprehensive comparative analysis of various approaches to automating camera trap image categorization. Our findings highlight the usefulness of integrating two independent classifiers—one specialized for cropped animals and the other for full images—with MegaDetector. This approach, where the appropriate classifier is selected based on whether the MegaDetector detects an animal, resulted in great improvements and allowed very limited overfitting to the location. 
More precisely, the combination of the MegaDetector with two fine-tuned BEiTV2 models reduced the relative error by around 42\%, 48\%, and 75\% on the CCT20, CEF22, and WCT test sets, respectively.

In addition, we proposed two alternative approaches for zero-shot classification based on (i) multi-modal models (e.g., BLIP and ChatGPT) and (ii) image retrieval based on deep features from DINOv2 and BioCLIP. For both, we used MegaDetector to detect and crop the animals.

Interestingly, DINOv2$_G$ achieved Top1 accuracy of 83.2\% and 87.5\% on CCT and CEF, respectively, trailing behind the best approach by 1.0\% and  4.7\% percentage points. This approach, leveraging robust visual representations and efficient similarity search, offers interesting properties such as (i) no requirement for fine-tuning while introducing new species and (ii) robustness toward location. 

The provided analysis underscores the value of using specialized classifiers tailored to different aspects of the image. When initial object detection fails, integrating a secondary classifier for full images substantially enhances the classification of images that MegaDetector deems empty, thus reducing overall classification errors. 
Besides, our results achieved in a zero-shot scenario showed a promising direction for zero-shot camera trap image classification methods. Future work should focus on developing those methods, as they allow for straightforward adaptability for new species and locations. \\

\noindent \textbf{The main outcomes derived from this work include the following:} \\

\noindent\textbf{MegaDetector rules.} 
The high accuracy and robustness in detecting animals, people, and vehicles in camera trap images make MegaDetector an excellent asset for preprocessing raw camera trapping data. Since CNN- and Transformer-based classifiers require resizing the original images to an expected input size, the detection and cropping of animals from the original images reduce the data loss related to resizing and subsequentially results in a considerable increase in performance and also a better regularization towards location changes. \\


\noindent\textbf{Large language and multi-modal models do not perform well.} While powerful in natural language processing and image captioning, ChatGPT and BLIP perform poorly on camera trap images. This is most likely due to their utility being confined to text- or text+image-based applications, making them unsuitable for the visual analysis of camera trap data. \\

\noindent\textbf{A great potential of retrieval-like classification.} DINOv2 features used in image retrieval settings for the classification of camera trap data are surprisingly robust, especially when using larger architectures. Comparing it with traditional classifier-based methods offers a promising alternative with similar performance and multiple benefits, e.g., new species or locations can be quickly and easily introduced just by adding a few samples into the \textit{database}, and no fine-tuning is ever needed, which saves a considerable amount of CO$_2$. \\

\vspace{-13pt}
\section*{Acknowledgements}
This research was supported by the Technology Agency of the Czech Republic, project No. SS05010008.
and by the grant of the University of West Bohemia, project No. SGS-2022-017.
Computational resources were
provided by the e-INFRA CZ project (ID:90254), supported by the Ministry of
Education, Youth and Sports of the Czech Republic. We also thank Friends of
the Earth, Czechia, and Národní Park Šumava for providing the data to form the CEF dataset.

%
%
\bibliographystyle{splncs04}
\bibliography{main}
\end{document}